\def\year{2022}\relax
\documentclass[letterpaper]{article} 
\usepackage{aaai22}  
\usepackage{times}  
\usepackage{helvet}  
\usepackage{courier}  
\usepackage[hyphens]{url}  
\usepackage{graphicx} 
\usepackage{natbib}  
\usepackage{caption} 
\DeclareCaptionStyle{ruled}{labelfont=normalfont,labelsep=colon,strut=off} 
\usepackage{algorithm}
\usepackage{algorithmic}
\usepackage{amsmath}

\usepackage{color}
\usepackage{graphicx}
\usepackage{paralist}
\usepackage{booktabs}

\usepackage{amsthm}
\newtheorem{definition}{Definition}
\newtheorem{theorem}{Theorem}
\newtheorem{example}{Example}

\let\oldexample\example
\renewcommand{\example}{\oldexample\normalfont}

\let\olddef\definition
\renewcommand{\definition}{\olddef\normalfont}

\let\oldtheorem\theorem
\renewcommand{\theorem}{\oldtheorem\normalfont}

\usepackage{newfloat}
\usepackage{listings}
\floatstyle{ruled}
\newfloat{listing}{tb}{lst}{}
\floatname{listing}{Listing}
\title{Activity Recognition in Assembly Tasks by Bayesian Filtering in Multi-Hypergraphs}

\author{
    Timon Felske$^1$, Stefan Lüdtke$^2$, Sebastian Bader$^1$, Thomas Kirste$^1$ 
}
\affiliations{
    $^1$ Institute of Visual \& Analytic Computing, University of Rostock, Germany\\
    $^2$ Institute for Enterprise Systems, University of Mannheim, Germany\\
    \{timon.felske, sebastian.bader, thomas.kirste\}@uni-rostock.de\\
    luedtke@es.uni-mannheim.de
}

\begin{document}

\maketitle

\begin{abstract}
We study sensor-based human activity recognition in manual work processes like assembly tasks. In such processes, the system states often have a rich structure, involving object properties and relations.
Thus, estimating the hidden system state from sensor observations by recursive Bayesian filtering can be very challenging, due to the combinatorial explosion in the number of system states.

To alleviate this problem, we propose an efficient Bayesian filtering model for such processes. In our approach, system states are represented by multi-hypergraphs, and the system dynamics is modeled by graph rewriting rules. 
We show a preliminary concept that allows to represent distributions over multi-hypergraphs more compactly than by full enumeration, and present an inference algorithm that works directly on this compact representation. 
We demonstrate the applicability of the algorithm on a real dataset.
\end{abstract}


\section{Introduction}


The automatic, sensor-based assessment of manual work processes is highly relevant in domains like  intralogistics \cite{reining2019human} or manufacturing \cite{tao2018worker}.
In this paper, we focus on \emph{assembly processes}, where a subject is assembling an object from multiple parts \cite{jones2021fine}. 
Tracking assembly processes can be used to assess process efficiency, to provide situation-aware assistance \cite{aehnelt2015information,gupta2012duplotrack}, or as the basis for human-robot interaction \cite{wang2020see}.

For these tasks, it is not sufficient to only estimate the current activity of the subject. Instead, we additionally need to estimate the current \emph{assembly state}  (also called \emph{context} \cite{ludtke2019human}), i.e., the state of all involved objects as well as their relations.
An established method for estimating hidden system states from a sequence of sensor data is \emph{recursive Bayesian filtering} \cite{sarkka2013bayesian}. In Bayesian filtering, a distribution $p(x_t | y_{1:t})$ over system states at time $t$ is estimated recursively, given the sequence $y_{1:t}$ of sensor data observed so far.
To see why Bayesian filtering in assembly processes can be difficult, consider the following example.


\begin{figure}
	\centering
	\includegraphics[width=0.4\textwidth]{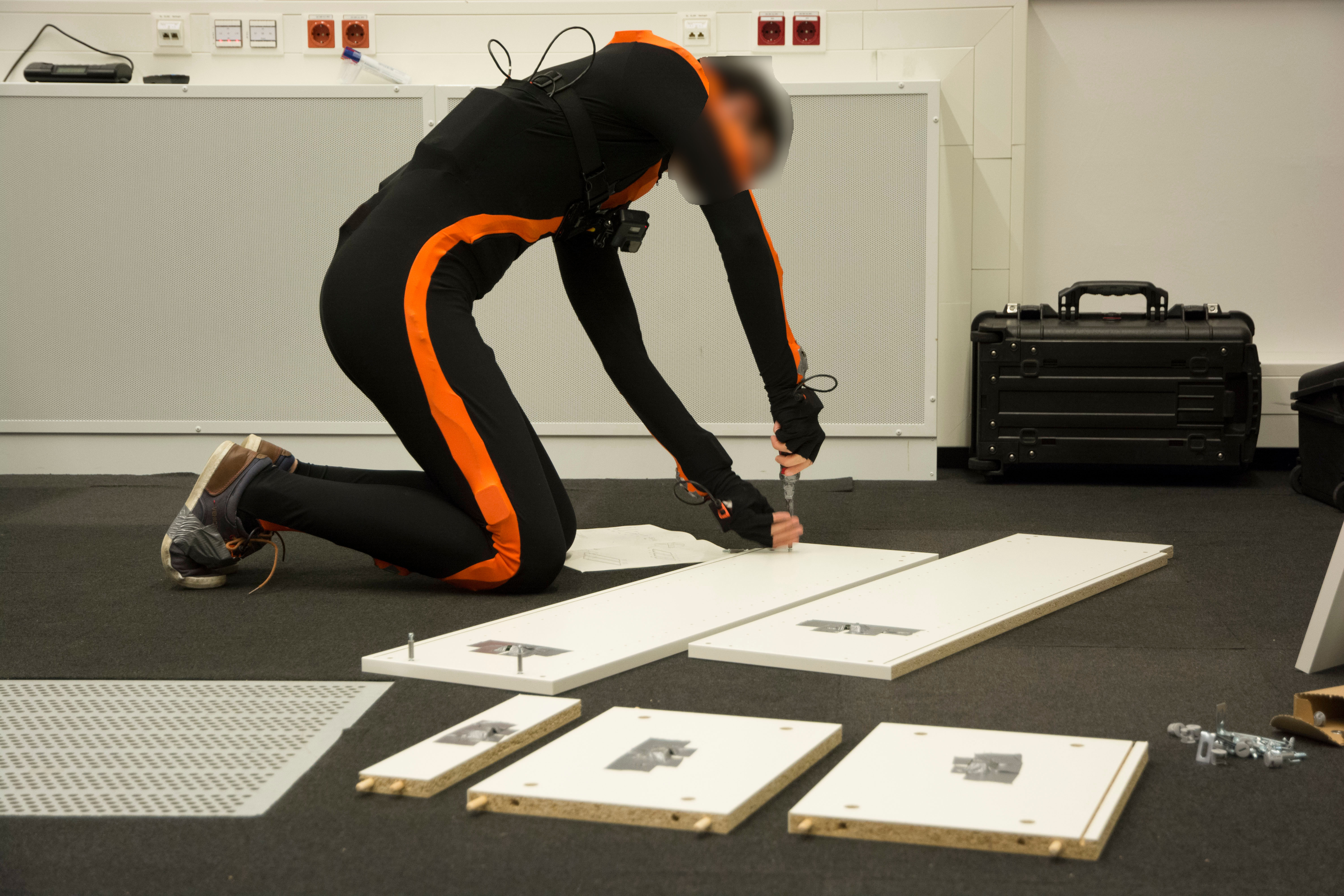}
	\caption{Participant assembling a bookcase, wearing a suit equipped with inertial sensors.}
	\label{Experiment}
\end{figure}

\begin{example}
A subject assembling a bookcase is instrumented with wearable sensors (see Figure \ref{Experiment}). Activities performed by the subject include, for example, picking up parts and tools, using tools, or connecting parts. A concrete example of an activity is the installation of an eccentric: The activity can only be applied when the subject holds both an eccentric and a screwdriver, results in the eccentric being attached to one of the boards.
\label{Example}
\end{example}

\begin{figure*}
	\centering
	\includegraphics[width=0.33\textwidth]{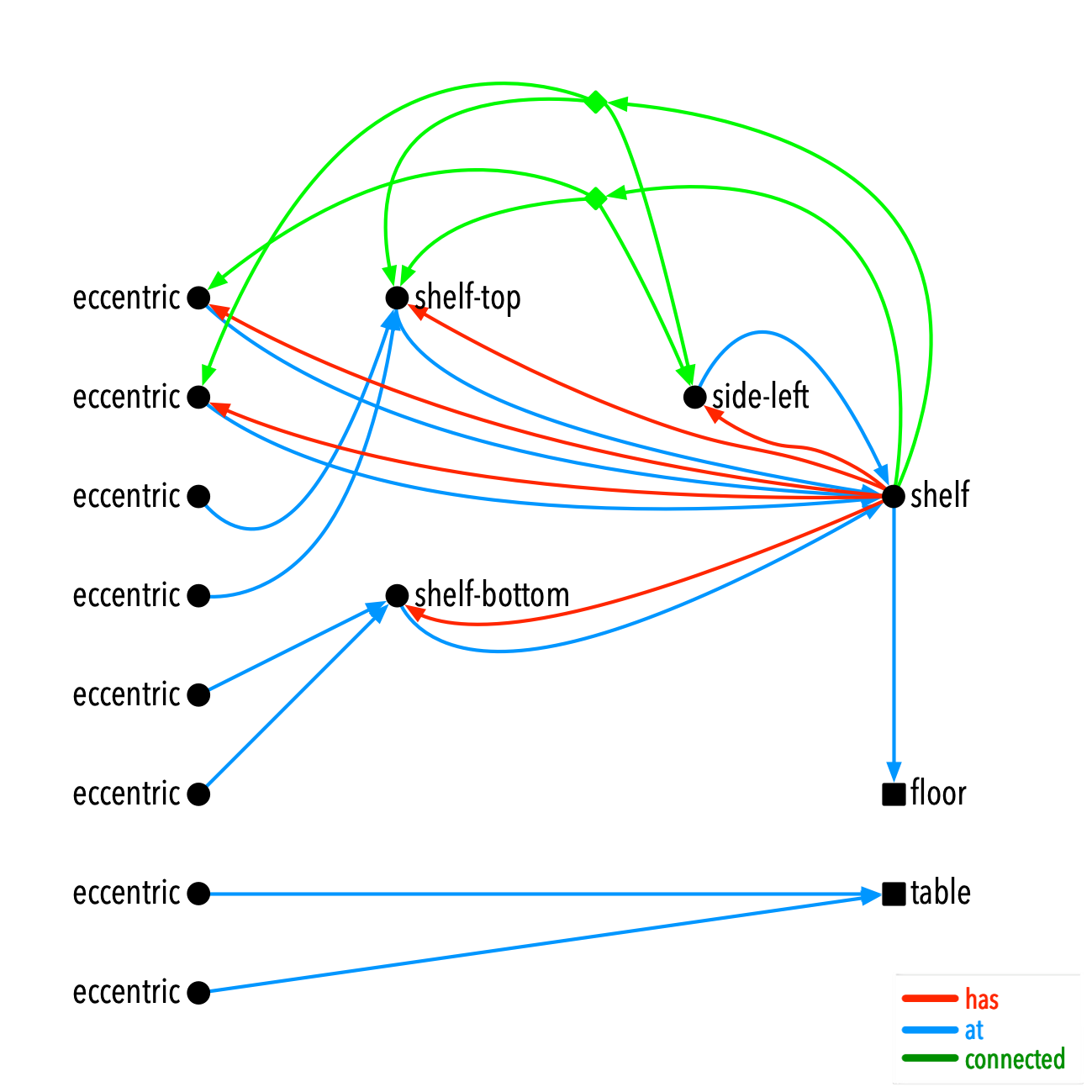}
	\includegraphics[width=0.33\textwidth]{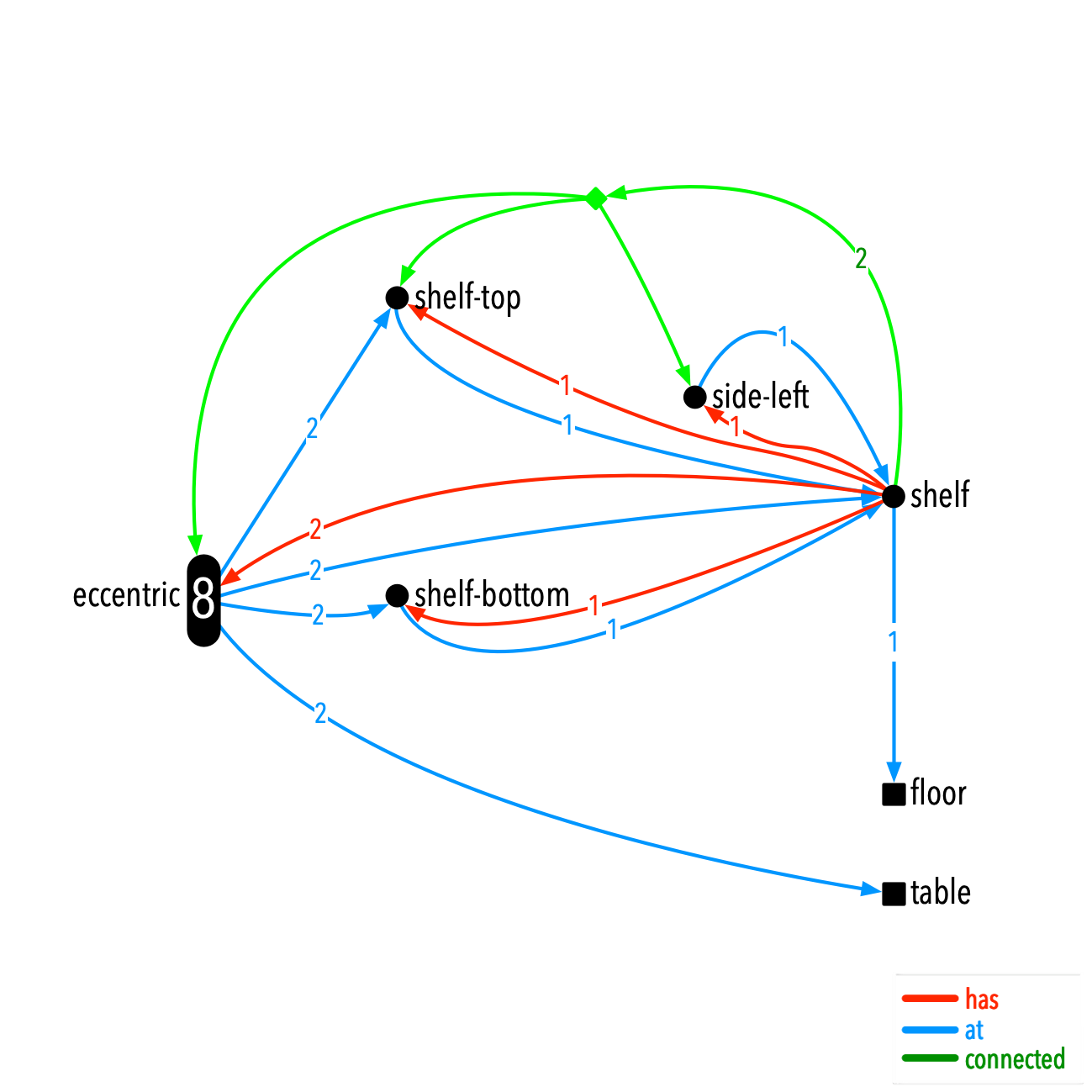}
	\includegraphics[width=0.33\textwidth]{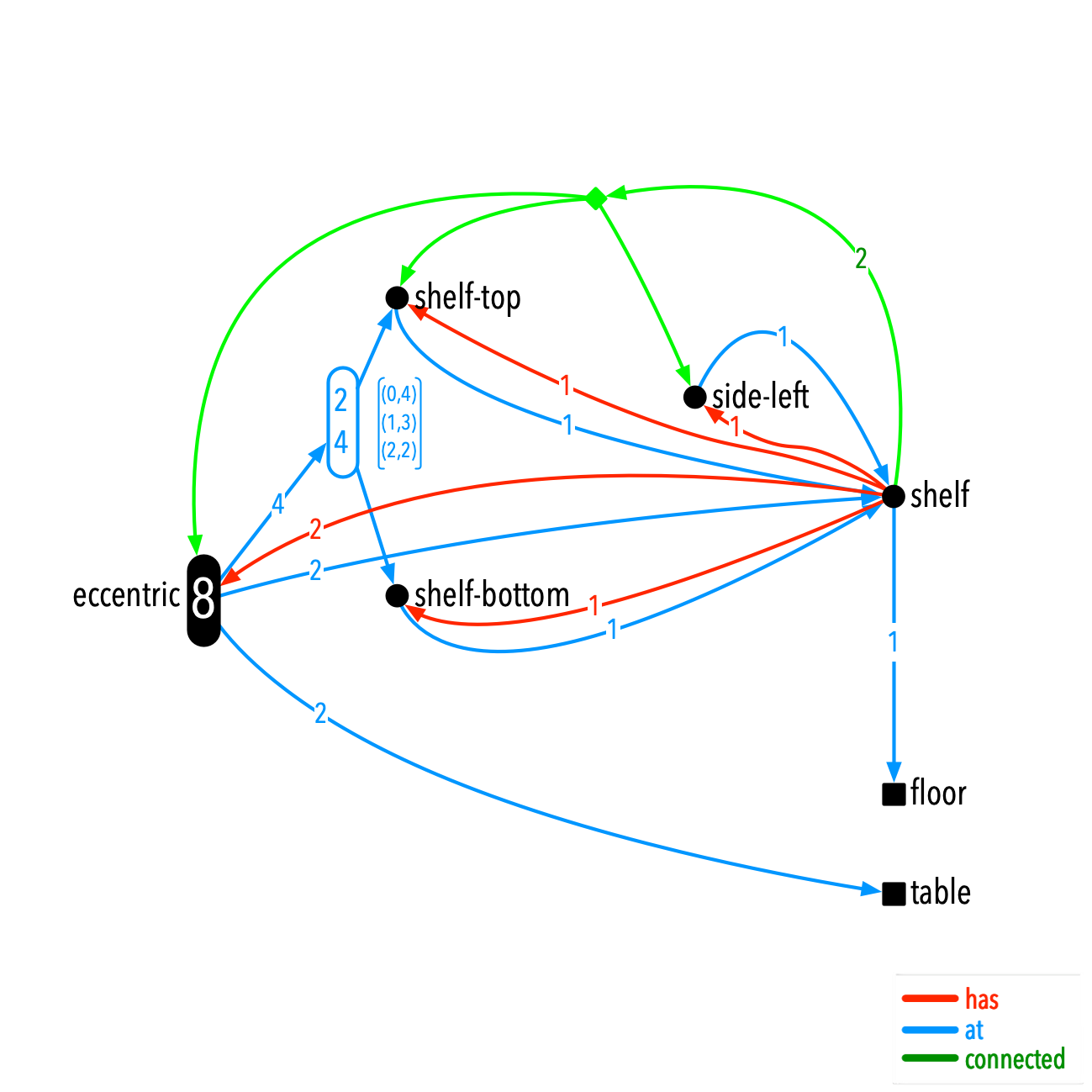}
	\caption{Graph-based representation of an assembly state. Left: hypergraph, Center: multi-hypergraph, Right: Lifted multi-hypergraph. The graphs describe the exact same state: Two eccentrics are located at the table, two are at the bottom-part of the shelf, two are at the shelf top and the upper two eccentrics are used to build a connection between the shelf-top and the side-left panel.}
	\label{MHGs}
\end{figure*}

The state of an assembly process usually consists of complex relations between objects, and the system dynamics can be described by rules that manipulate these relations. 
Such assembly processes can be modeled as \emph{graph rewriting systems}, where system states are edge- and node-labelled graphs, and the system dynamics is represented by graph rewriting rules \cite{jones2021fine}.
\citet{wang2016weighted} also used weighted graphs to represent different sequences to assemble a product and find the optimal assembly sequence.

Unfortunately, Bayesian filtering in such systems can be extremely challenging due to the large number of discrete system states, arising from the combinatorial explosion in the properties of individual objects and their relations. 

Therefore, methods for efficiently representing distributions over multi-hypergraphs, and efficient inference algorithms are essential for sensor-based assembly tracking. In this paper, we report on our ongoing work in that direction. Specifically, this paper has the following contributions:
\begin{itemize}
    \item We introduce the general concept of Bayesian filtering in multi-hypergraph rewriting systems.
    \item We propose a representation of distributions over multi-hypergraphs that can be much more efficient than enumerating all graphs with non-zero probabilty. The representation exploits \emph{symmetries} in the distribution, arising from the fact that not all parts need (or can) be distinguished. 
    \item We present an efficient Bayesian filtering algorithm that works directly on this compact representation.
\end{itemize}
Finally, we outline a number of challenges and directions for future work.


\section{Example Domain: Bookshelf Assembly}
As a concrete example, we focus on the assembly of a bookshelf, as already introduced above. Here, we describe the domain in more detail to highlight algorithmic challenges in this domain.

This assembly task consists of 56 different components (entities).
The entities can be divided into three categories: 7 boards, 40 screws and 9 tools.
Of these individual entities, the screws are indistinguishable from one another.
In our scenario, a single agent is assembling the bookshelf. 
While doing that, the agent wears a suit equipped with inertial measurement units (IMUs), containing accelerometer and gyroscope sensors. 
Like the agent, the various entities are also equipped with IMUs.

Furthermore, each step of the assembly is annotated offline based on video recordings of the assembly.
Annotations are 5-tuples $(a_t,l_t,l_{t+1},o_t,o_{t+1})$, where $a_t$ is the action class at time $t$, $l_t$ and $l_{t+1}$ are the locations at times $t$ and $t+1$, and $o_t$ and $o_{t+1}$ are objects held by the agent at times $t$ and $t+1$. 
In case multiple objects are held, the amount of held objects is given.
For example, the annotation where an agent takes an eccentric from the floor is represented by the 5-tuple $(take, floor, floor, (eccentric\; 1), (eccentric\; 2))$.

As the focus of this paper is the representation of states and system dynamics but not sensor models, we assume that annotation sequences can be observed directly. The observation model is defined such that $p(y_{t+1}|x_{t+1},a_t,x_t) = 1$ if the system states $x_t$ and $x_{t+1}$ and action $a_t$ are consistent with the observation (i.e., annotation) $y_t$ and $0$ otherwise. 
Investigating more realistic observation models that involve the IMU data is a topic for future work. 


\section{Multi-Hypergraph States}

Since we want to apply a Bayesian filtering algorithm on our Bookshelf-Assembly task, we need to represent the different states of the assembly.
In the domain of assembly tasks, graphs are often used to represent the state of the assembly \cite{jones2021fine,wang2016weighted}. 
With this data structure, it is possible to efficiently represent the entities and their relation to one another.

Relations can involve more than two entities, e.g., two boards being connected by an eccentric. \emph{Hypergraphs}, can naturally represent these cases. In a hypergraph, edges can connect more than two vertices. Formally, a hypergraph is a pair $(V,E)$, where $V$ is a set of vertices and the set of hyperedges $E \subseteq \mathcal{P}(V)$, where $\mathcal{P}(V)$ is the power set of $V$. 
An example of a hypergraph for the bookshelf domain is shown in Figure \ref{MHGs} (left).

As mentioned before, we can not distinguish between the individual screws (eccentric, etc.).
Therefore, the hypergraph in Figure \ref{MHGs} can be improved further: Instead of representing the indistinguishable entities as individual vertices in the graph, we can use a \emph{multigraph}. Formally, a multigraph is a pair $(V,E)$, where $V$ is a set of vertices and $E$ a set of edges.
Furthermore, the vertices and edges have associated \emph{multiplicities}. 
The summed multiplicity of all edges connected to each vertex needs to be equal to the multiplicity of that vertex.
Thus, overall, an assembly state can be represented by a \emph{multi-hypergraph} (MHG), as shown in Figure \ref{MHGs} (center).
These multiplicities are also instrumental for more efficient representations of \emph{distributions} over MHGs, as shown below.

To apply Bayesian Filtering to MHGs, we need to model the system dynamics w.r.t. these graphs.
We use a graph rewriting formalism for this, as discussed next.

\section{Bayesian Filtering in Multi-Hypergraphs}
In this chapter, we describe how to apply Bayesian Filtering to multi-hypergraphs.

We use \emph{graph rewriting rules} to specify the system dynamics. This way, a model of the system dynamics can be constructed from prior domain knowledge, instead of learning it from data. For example, in the bookshelf assembly domain, rewriting rules that describe how parts can be connected can be derived directly from a construction manual. 
The knowledge-based construction of transition models is particularly advantageous when only a small amount of training data is available compared to the number of possible activity trajectories---as usual in human activity recognition.
Furthermore, preconditions of rules can be used to reduce the set of possible actions based on the current state and the observations while filtering.

\begin{figure}
	\centering
	\includegraphics[width=0.33\textwidth]{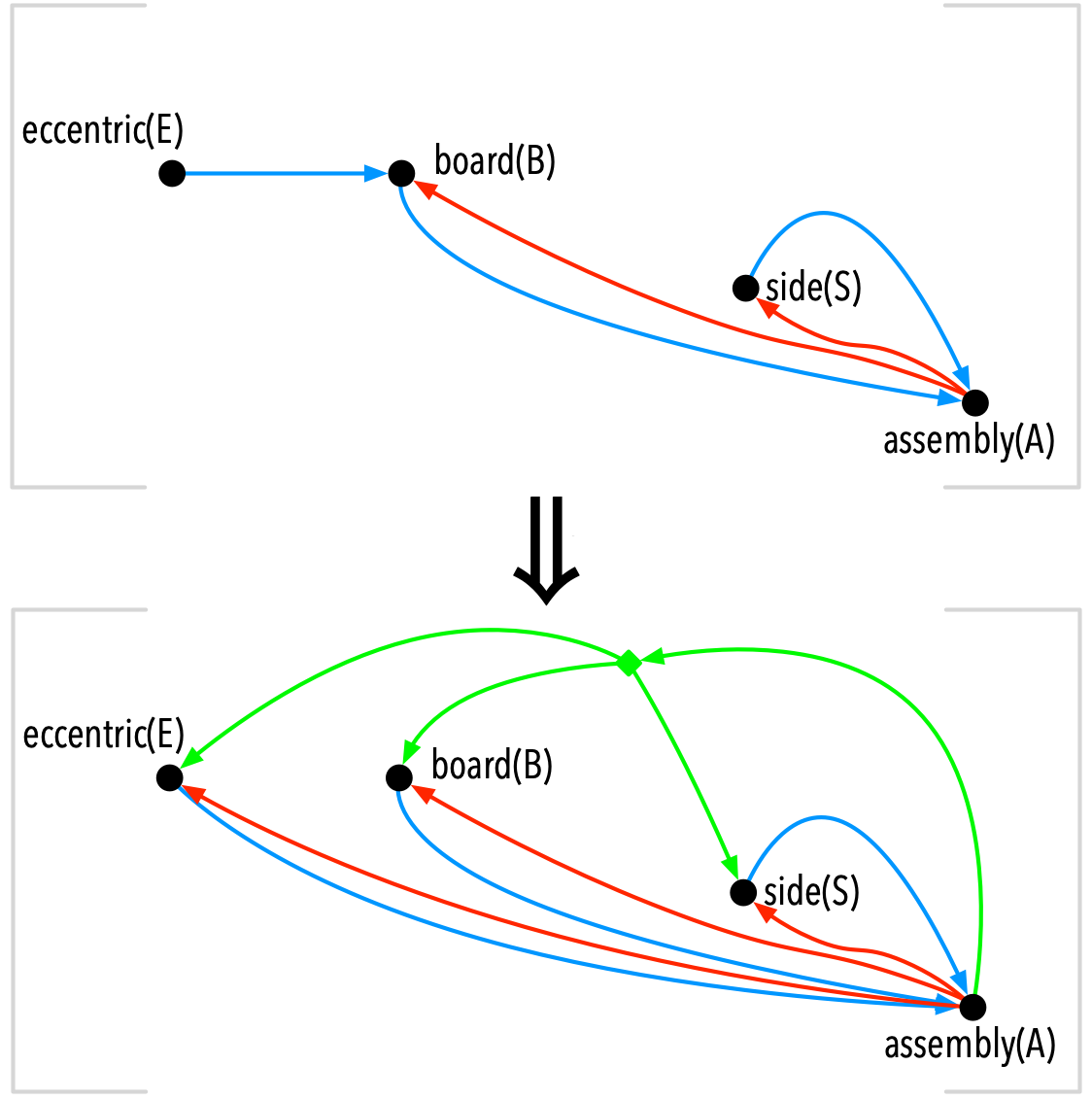}
	\caption{Graphical representation of the described rewriting rule \textit{installEccentric}. The upper graph pattern needs to exist in the state for the rule to be applicable. The rule transforms that pattern to the lower pattern.}
	\label{rewriting}
\end{figure}

As an example of a graph rewriting rule, consider the rule shown in Figure \ref{rewriting}. The rule consists of a precondition (a graph pattern that needs to exist in the state for the rule to be appliable) and an effect (which describes how the sub-graph corresponding to the precondition is changed when the rule is applied). 
Specifically, the rule \textit{installEccentric} retracts the \textit{at}-edge between an eccentric and its current location.
After that, a new edge between the eccentric and its new location (where it is installed) is created.
Furthermore the \textit{has}-edge between these two entities will be added.
Last, the \textit{connected}-edge between the eccentric and the connected entities of the bookshelf will be realised.

Graph rewriting systems on multi-hypergraphs are a generalization of multiset rewriting systems, as used in Lifted Marginal Filtering \cite{ludtke_lifted_2018}.
Specifically, from the viewpoint of multiset rewriting systems, graph patterns are non-local preconditions (constraints), involving agreement of values of different entities in the multiset. Such constraints cannot be modeled in Lifted Marginal Filtering due to the simple constraint language which is used to guarantee that constraint satisfaction is tractable. Instead, to test graph pattern constraints, a more general approach like lifted weighted model counting \cite{gogate2011probabilistic} (which can test constraints without grounding the model completely) could be required.

\begin{figure}
	\centering
	\includegraphics[width=0.2\textwidth]{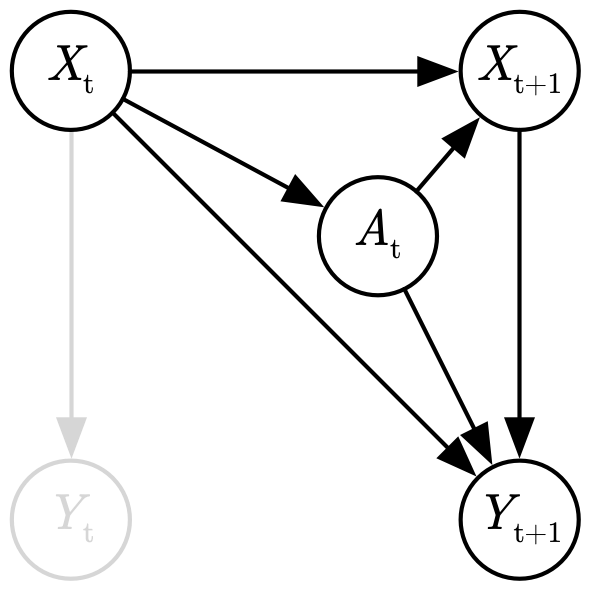}
	\caption{Graphical model representation of the Bayesian filtering model described by our approach. $A_t$ are rewriting rules, $X_t$ and $X_{t+1}$ are multi-hypergraph states, and $Y_{t}$ and $Y_{t+1}$ are observations.}
	\label{action-observations}
\end{figure}

As illustrated in Figure \ref{action-observations}, the transition model is given by
\begin{align}
p(x_{t+1} \mid x_t) = \sum_{a_t} p(x_{t+1} \mid a_t, x_t)\: p(a_t\mid x_t).
\end{align}
Here, $a_t$ is a rewriting rule, the distribution $p(x_{t+1} \mid a_t, x_t)$ specifies the states $x_{t+1}$ that result from applying rule $a_t$ to state $x_t$ and $p(a_t \mid x_t)$ is the participants' action selection model. 

The observation model is given by $p(y_{t+1}\mid x_{t+1},a_t,x_t)$.
This represents the idea that observations reflect what happens during the interval $(t,t+1]$, depending on the action $a_t$ as well as on the states $x_t$ and $x_{t+1}$ present before and after this action.

For recursive Bayesian filtering, we are interested in recursively estimating the marginal filtering distribution $p(x_{t+1} \mid y_{1:t+1})$ at time $t+1$ from the filtering distribution at time $t$, the transition model and the observation model. In principle, we can use the usual Bayesian filtering prediction and update equations, but need to account for the action variable $A_t$ and the fact that the observation model depends on $A_t$ and $X_{t+1}$. This way, the prediction becomes
\begin{align}
\begin{split}
& p(x_t,a_t,x_{t+1} \mid y_{1:t}) 
= \\ & p(x_{t+1} \mid a_t,x_t)\: p(a_t \mid x_t) \: p(x_t | y_{1:t})
\end{split}
\end{align}
and the update is computed as
\begin{align}
\begin{split}
& p(x_{t+1} \mid y_{1:t+1})
= \\ & \frac 1 Z \sum_{a_t,x_t} p(x_t,a_t,x_{t+1} \mid y_{1:t}) \: p(y_{t+1}\mid x_{t+1},a_t,x_t),
\end{split}
\end{align}
where $\frac 1 Z$ is a normalization factor. 
Note that marginalization requires to evaluate whether  states are identical, i.e., solve a graph isomorphism problem. Thus, future work needs to focus on special cases where graph isomorphism can be solved efficiently, e.g.\ via appropriate graph canonization.

In general, the distribution $p(x_t \mid y_{1:t})$ can have very many states with non-zero probability. For example, when there are $k$ screws and $n$ holes, there are $\binom{n}{k}$ ways to attach the screws to holes. Thus, we are interested in efficient representations of such distributions, which do not rely on full enumeration. 

We propose \emph{lifted multi-hypergraphs} to efficiently represent distributions over multi-hypergraphs (LMHGs). Each LMHG represents a \emph{distribution} over (ground) MHG. Conceptually, LMHGs are an extension of \emph{lifted multiset states}, as used in Lifted Marginal Filtering \cite{ludtke_lifted_2018}. 
The illustrated example represents the case where 4 eccentrics are attached to the \textit{shelf-top} and \textit{shelf-bottom} boards.
At most two of them are at \textit{shelf-top} and at most 4 of them are at \textit{shelf-bottom}.
This results in three different situations of how eccentrics could be distributed. The LMHG represents a \emph{uniform} distribution over these situations.

Since we apply Bayesian filtering to LMHGs, the rewriting rules need to be adapted in order to implement the system dynamics correctly.
More precisely, we need rules that describe the extent to which the distributions change when indistinguishable entities are installed at unknown locations.
To stick with the example discussed above:
Suppose that an observation indicates that an eccentric is installed, but the exact location of the eccentric is unknown.
According to the specified rule, the total amount of installed eccentrics is increased by 1, and the maximum amount of installed eccentrics at each reachable location is increased by 1. 
To maintain the integrity constraint, the count of installed eccentrics at any location can not be larger than the total count of installed eccentrics.
Applying this rule to LMHGs can be understood as applying a grounded version of the rule to each specific MHG that is contained in the LMHG. 

In contrast to \citet{ludtke_lifted_2018}, we assume that similar parts never need to be distinguished explicitly, thus we do not require a \emph{splitting} operator that would handle identification.

\section{Experimental Evaluation}

\begin{table}[]
    \centering
    \begin{tabular}{cccc}
    \toprule
        Task & N & duration (min) & actions \\
        \midrule
        Normal & 11 & $12.3 \pm 5.1$ & $360.5 \pm 127.3$ \\ 
        Error & 12 & $7.8 \pm 5.8$ & $224.7 \pm 147.3$\\
        \bottomrule
    \end{tabular}
    \caption{Features extracted from the recording of 11 normal and 12 (deliberately) erroneous experiments.}
    \label{tab:experimentalData}
\end{table}



In this section, we demonstrate the general applicability of our concept to a real dataset.
The dataset was created by recording sensor data of subjects assembling a bookshelf, as introduced above. 
Subjects wore a body suit with 17 IMUs, and all objects (except screws) were equipped with IMUs as well. All experiments were recorded on video for offline annotation. 

Overall, we performed 23 experiments with 12 different subjects.
Each subject was supposed to do a successful (correct) and an erroneous bookcase assembly. The erroneous runs were recorded as one of our future research goals is to \emph{detect} assembly errors. 
We intended to perform 24 experiments to generate data for 12 successful and 12 erroneous assemblies. The data of one successful assembly was not usable, resulting in 11 included erroneous experiments.
The 23 experiments provided recordings with 240.9 minutes of relevant data.
Properties of this data is listed in Table \ref{tab:experimentalData}.

Currently, we concentrate on the evaluation based on the annotations to demonstrate the basic applicability of our approach.
Our filtering model was able to explain all 11 correct annotation sequences. During filtering, at most 2 lifted multi-hypergraphs were required to represent the marginal filtering distribution for all sequences and all timesteps, instead of approx. 5000 in the grounded version.
The significant reduction of the necessary states can be explained by the use of LMHGs, which allow to  represent several states by a single representative.
This initial experiment shows that our Bayesian filtering approach can be applied to track assembly processes and efficiently represent the filtering distribution.

\section{Discussion and Conclusion}
In this paper, we presented a Bayesian filtering model with multi-hypergraph states and graph rewriting-based system dynamics. 
The main technical contribution is an efficient representation of distributions over multi-hypergraphs and a Bayesian filtering algorithm that works directly on that representation.  
Our approach was motivated by state estimation in assembly processes. However, the method can be usefully employed to other state estimation tasks in a dynamic system that consists of multiple entities and their relations, and where the system dynamics is naturally described by rewriting rules, e.g. multi-agent systems or social networks. 

To make our approach applicable to real-world domains, several extensions are required: 
First, we did not discuss the \emph{observation model} $p(y_t | x_t)$ here, which relates sensor data to system states. Apart from simple, parametric densities, generative neural networks (e.g. normalizing flows \cite{rezende2015variational}) could be employed. 
Second, in real-world datasets, actions have distinct \emph{durations}, that need to be modeled appropriately, similar to methods used for hidden semi-Markov models.
Third, our future work will focus on more general means to efficiently represent distributions over graphs, as well as a formal analysis of the expressiveness and computational complexity of our approach.



\section{Acknowledgments}
This work was funded by the European Social Fund (ESF) and the Ministry of Education, Science and Culture of Mecklenburg-Western Pomerania (Germany) within the project NEISS – Neural Extraction of Information, Structure and Symmetry in Images under grant no ESF/14-BM-A55-0009/19.


\bibliography{aaai22}

\end{document}